\documentclass[APA,LATO1COL]{WileyNJD-v2}

\articletype{Article Type}%

\received{8 July 2021}
\revised{6 June 2016}
\accepted{6 June 2016}

\usepackage{amsfonts} 
\usepackage{subcaption} 

\begin{document}

\title{Epsilon Consistent Mixup: Structural Regularization with an Adaptive Consistency-Interpolation Tradeoff}

\author[1]{Vincent Pisztora*}

\author[2]{Yanglan Ou}

\author[2]{Xiaolei Huang}

\author[1,3]{Francesca Chiaromonte}

\author[1]{Jia Li}

\authormark{PISZTORA \textsc{et al}}

\address[1]{\orgdiv{Department of Statistics}, \orgname{Pennsylvania State University}, \orgaddress{\state{PA}, \country{USA}}}

\address[2]{\orgdiv{Department of Information Sciences and Technology}, \orgname{Pennsylvania State University}, \orgaddress{\state{PA}, \country{USA}}}

\address[3]{\orgdiv{EMbeDS}, \orgname{Sant'Anna School of Advanced Studies}, \orgaddress{\state{Pisa}, \country{Italy}}}

\corres{*Vincent Pisztora, Department of Statistics, Pennsylvania State University, State College, PA, USA. \email{uxp5@psu.edu}}

\presentaddress{Department of Statistics, Pennsylvania State University, State College, PA, USA. \email{uxp5@psu.edu}}

\abstract[Summary]{In this paper we propose Epsilon Consistent Mixup ($\epsilon$mu). $\epsilon$mu is a data-based structural regularization technique that combines Mixup's linear interpolation with consistency regularization in the Mixup direction, by compelling a simple adaptive tradeoff between the two. This learnable combination of consistency and interpolation induces a more flexible structure on the evolution of the response across the feature space and is shown to improve semi-supervised classification accuracy on the SVHN and CIFAR10 benchmark datasets, yielding the largest gains in the most challenging low label-availability scenarios. Empirical studies comparing $\epsilon$mu and Mixup are presented and provide insight into the mechanisms behind $\epsilon$mu's effectiveness. In particular, $\epsilon$mu is found to produce more accurate synthetic labels and more confident predictions than Mixup.
}

\keywords{Consistency Regularization, Data Augmentation, Epsilon Consistent Mixup, Mixup, Semi Supervised Learning, Structural Regularization}

\jnlcitation{\cname{%
\author{Pisztora V.}, 
\author{Ou Y.}, 
\author{Huang X.}, 
\author{Chiaromonte F.}, and 
\author{Li J.}} (\cyear{2021}), 
\ctitle{Epsilon Consistent Mixup: Structural Regularization with an Adaptive Consistency-Interpolation Tradeoff}, \cjournal{Stat}, \cvol{2021;00:1--13}.}

\maketitle

\section{Introduction}
\label{intro}

An increasingly important challenge facing modern statistical learning is effective regularization. As the field has tackled progressively more ambitious problems, the models needed to learn such tasks have necessarily become more powerful and thus more susceptible to overfitting. In many applications however, the amount of labelled data needed to support generalization of these models is rarely available, even as unlabeled data is often plentiful. In such cases, regularization methods are needed to reduce overfitting by imposing additional constraints on learning \citep{a1_Goodfellow_book_2016}. Regularization methods able to operate in a semi-supervised fashion, utilizing both labeled and unlabeled observations, have been particularly effective \citep{113_meanteach} \citep{111_Miyato2019VirtualAT} \citep{112_Verma2019ICT} \citep{46_fxmatch}. 

In general, regularization procedures discourage selection of overfitted models by biasing selection against various forms of model complexity. To induce this preference for simpler models, these methods modify various elements of the learning process including the model specifications themselves \citep{105_dropout} \citep{108_Ioffe2015BatchNA} \citep{106_Huangstochdepth} \citep{107_Gastaldi2017ShakeShakeRO}, the optimization algorithm \citep{109_Krogh1991_weightdecay} \citep{110_Caruana2000OverfittingIN}, and the training data \citep{1_zhang2018mixup} \citep{111_Miyato2019VirtualAT} \citep{3_2019_manifoldmixup} \citep{77_Shorten2019ASO} \citep{12_Yun2019CutMixRS} \citep{15_Summers2019ImprovedMD}. The focus of this work is on the latter data-based procedures which we name structural regularization methods. 

\begin{figure}[tb]
    \centering
    \includegraphics[scale=0.25]{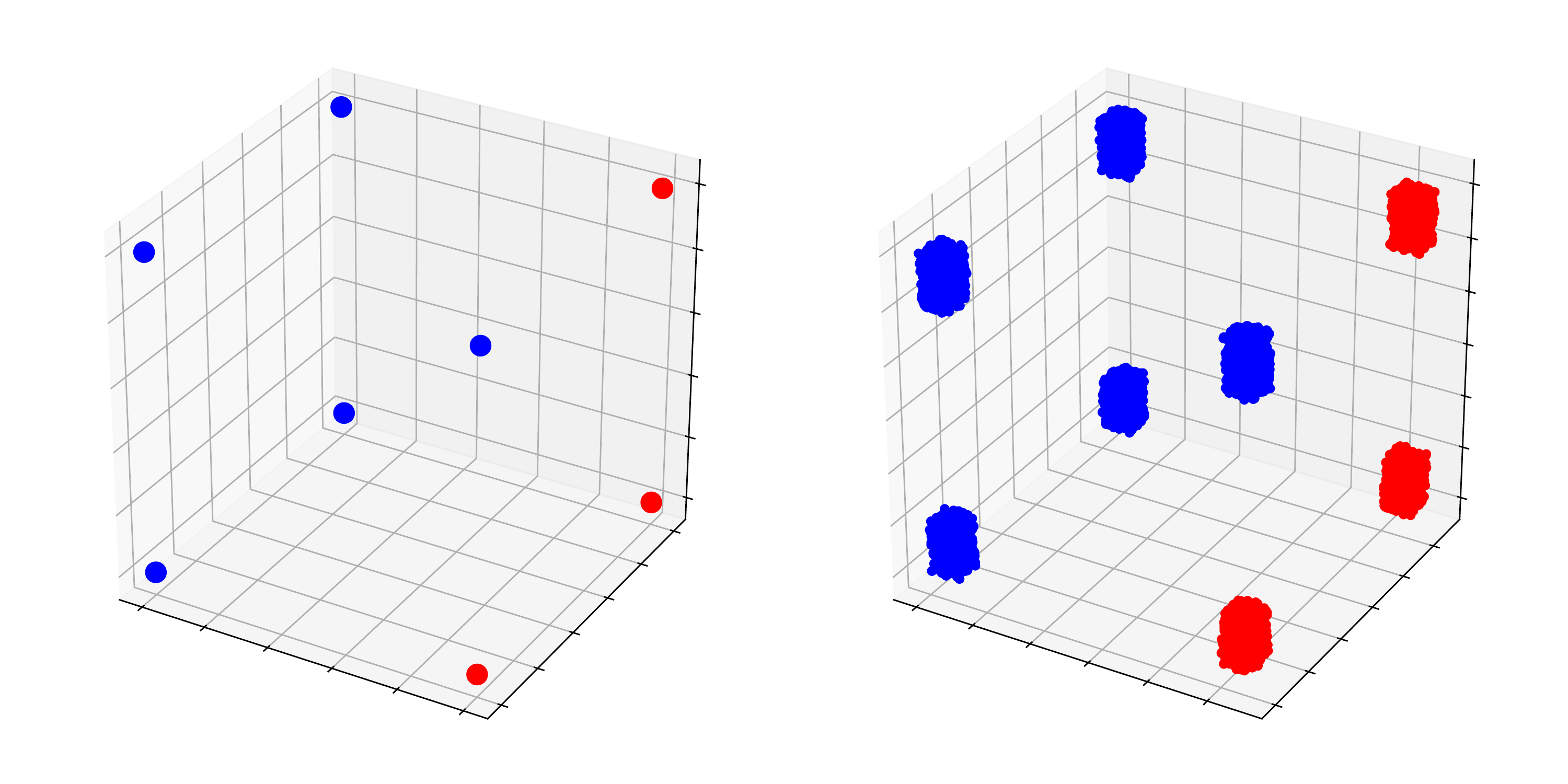}
    \caption{This figure illustrates how feature and response transformations ($T_\mathcal{X}$,$T_\mathcal{Y}$) can be thought of as generating structures in the feature space. In this case the plot on the left shows the fully labelled data $D^N$ in a 3D feature space with color representing class. The plot on the right shows the synthetic data $\tilde{D}^{\tilde{N}}$ generated by repeatedly applying uniform noise to each $x\in X^N$ unevenly across dimensions using $T_\mathcal{X}$, while maintaining the class labels locally around each original observation using $T_\mathcal{Y}$ - thus producing rectangular structures of the same class as the nearest original observation $x$. Notice in this case $N<\tilde{N}$.}
    \label{fig:3dstruc}
\end{figure}

Structural regularization procedures work by transforming observed datasets to either introduce new, or amplify existing, desired auxiliary relationships between features and responses. As demonstrated in Fig. \ref{fig:3dstruc}, these relationships can be thought of as representing geometric structures in the feature space. Inclusion of these relationships in the transformed synthetic dataset indirectly restricts the set of viable candidate models by increasing the specificity of model requirements and thus produces a regularizing effect. 

Training using these synthetic datasets incentivizes selection of models preserving the added relationships built into the data via transformation. As an example, class invariance to image rotation can be obtained by replacing the observed data with rotations of images paired with the labels of the unrotated originals. Models trained on this synthetic dataset learn to ignore rotation in making classification decisions, thus incorporating this rotation invariance into the final classifier.

In this paper we introduce a new structural regularization method, Epsilon Consistent Mixup ($\epsilon$mu). This method has three distinct elements of novelty. First, $\epsilon$mu is shown to outperform standard Mixup in semi-supervised classification accuracy, synthetic label quality, and prediction confidence. Second, $\epsilon$mu successfully combines two classes of structural regularization, namely consistency and interpolation. Finally, $\epsilon$mu achieves this combination through an adaptive mechanism that allows the optimal synthetic structure, a balance between consistency and interpolation, to be learned during training using simple gradient-based optimization.

We organize this paper as follows. In Section 2, we first introduce the general structural regularization framework. We then describe two special cases, consistency regularization and Mixup interpolation, which are components to the proposed Epsilon Consistent Mixup ($\epsilon$mu). In Section 3.1, we introduce our proposed structural regularization method, $\epsilon$mu. In Section 3.2, we present experimental results demonstrating $\epsilon$mu's advantages in classification performance, synthetic label quality, and prediction entropy.

\section{Structural Regularization}
\label{struc_reg}

In this section we first define the semi-supervised setting. We then introduce the general formulation for structural regularization in this setting. Finally, we describe two special cases of structural regularization, consistency regularization and Mixup interpolation, which are key components of the proposed Epsilon Consistent Mixup.

\subsection{Semi-Supervised Setting}

We consider the general semi-supervised setting, defining a dataset $D^N=\{D_L^{n_L},D_U^{n_U}\}$ where $N=n_U+n_L$, $D_L^{n_L}=(X_L^{n_L},Y_L^{n_L})=\{(x_i,y_i):i=1,...,n_L\}$ represents the set of $n_L$ labelled observations, and $D_U^{n_U}=(X_U^{n_U},Y_U^{n_U})=\{(x_i,y_i) : y_i = \emptyset ~ \forall i,i=n_L+1,...,N\}$ represents the set of $n_U$ unlabeled observations, with the placeholder $y_i=\emptyset$ standing in for the missing labels of the unlabeled observations. 
We take each observation's features to be $x\in\mathcal{X}$ and a data matrix of $n$ observations' features to be $X^n\in\mathcal{X}^n$, where $\mathcal{X}^n$ is the $n$-fold Cartesian product of $\mathcal{X}$. Each observation's response is taken to be $y\in\mathcal{Y}$ with a response matrix of $n$ such responses being $Y^n\in\mathcal{Y}^n$ where $\mathcal{Y}^n$ is the $n$-fold Cartesian product of $\mathcal{Y}$. Then dataset $D^n\in (\mathcal{X}\times\mathcal{Y})^n = \mathcal{D}^n$
where $(\mathcal{X}\times\mathcal{Y})^n$ is the $n$-fold Cartesian product of $\mathcal{X}\times\mathcal{Y}$, itself the Cartesian product of $\mathcal{X}$ and $\mathcal{Y}$.
We also take $x\sim p (x)$, and $(x,y)\sim p(x,y)$. 

In the analysis that follows, we consider the semi-supervised image classification task in this setting. We take $\mathcal{Y}=\mathbb{R}^C\cup\emptyset$, where each response $y\in \mathcal{Y}$ is either unlabeled, and represented by the placeholder $y=\emptyset$, or is represented as the conditional probability of membership in each of $C$ possible classes ($y\in \mathbb{R}^C$). We also take $\mathcal{X}=\mathbb{R}^{h\times w \times c}$ where $h,w,c\in \mathbb{Z}$ represent, respectively, the height, the width, and the number of color channels of each image. Our objective is to estimate the conditional class probability $p(y\vert x)$ with the neural network estimator $f_\theta (x)\in \mathcal{F}$
parameterized by weights $\theta \in \Theta$.

\subsection{General Formulation}

Structural regularization is a general formulation of data-driven regularization schemes that provides a useful framework for comparing individual methods. In this formulation, each data-driven regularization method is defined as a specific triplet $(P,T,L_S)$, of pseudo-label function, transformation function, and structural loss function.

The implementation of structural regularization methods begins with the application of the pseudo-label function $P:(\mathcal{D}^{n_U},\mathcal{F})\to \mathcal{D}^{n_U}$ to the unlabelled observations $D^{n_U}_U$, producing the pseudo labelled dataset $\hat{D}^N=\{D^{n_L}_L,\hat{D}^{n_U}_U\}$ where $\hat{D}^{n_U}_U = (X^{n_U}_U,\hat{Y}^{n_U}_U)= P(D^{n_U}_U,f_{\hat{\theta}})$ and $f_{\hat{\theta}}$ is an estimated classifier. This pseudo-labelling step enables the use of structural regularization in the semi-supervised setting by generating a pseudo-label $\hat{y}\in \hat{Y}^{n_U}_U$ for each unlabelled observation $(x,y)\in D^{n_U}_U$.

The transformation function $T: \mathcal{D}^N\to \mathcal{D}^{\tilde{N}}$ then generates the synthetic dataset $\tilde{D}^{\tilde{N}}=(\tilde{X}^{\tilde{N}},\tilde{Y}^{\tilde{N}})$ by either introducing new, or amplifying existing, desired relationships between covariates and responses into the pseudo-labelled dataset $\hat{D}^N$ (as in Fig. \ref{fig:3dstruc} for example). The function $T$ is comprised of two components, $T_\mathcal{X} : \mathcal{D}^N\to \mathcal{X}^{\tilde{N}}$, responsible for generating the features of $\tilde{N}$ observations in $\tilde{D}^{\tilde{N}}$, and $T_\mathcal{Y} : \mathcal{D}^N\to \mathcal{Y}^{\tilde{N}}$, responsible for generating the respective responses $\tilde{Y}^{\tilde{N}}$. 

Preference for the auxiliary relationships encoded in $\tilde{D}^{\tilde{N}}$ is then stipulated with an additional structural loss term $L_S$ (Eq. \ref{eq:2}) in the total loss function.
\begin{equation}\label{eq:2}
    L_S = \sum_{(\tilde{x}_i,\tilde{y}_i) \in (\tilde{X}^{\tilde{N}},\tilde{Y}^{\tilde{N}})} R(f_{\theta}(\tilde{x}_i),\tilde{y}_i)
\end{equation}
This structural loss penalizes discrepancy between the predicted conditional probability $f_{\theta}(\tilde{x}_i)$ and the desired class probabilities $\tilde{y}_i$, as measured by dissimilarity measure $R$ across the observations $(\tilde{X}^{\tilde{N}},\tilde{Y}^{\tilde{N}})$. Thus, the preservation of the relationships built into $\tilde{D}^{\tilde{N}}$ through $T$ is incentivized in the eventual fitted classifier by this additional loss.

Model training is then performed using the total loss function $L_T$ (Eq. \ref{eq:LT}) which is a sum of the supervised loss $L$ and the structural loss $L_S$ weighted by $w_S$, the structural loss weight. The supervised loss operates on a subset $\tilde{D}^{\tilde{n}_L}_{L}$ of the synthetic dataset, which can be thought of as corresponding to the originally observed labelled data $D^{n_L}_{L}$.
\begin{equation}\label{eq:LT}
    L_T(f_{\theta},\tilde{D}^{\tilde{N}})=L(f_{\theta},\tilde{D}^{\tilde{n}_L}_{L})+w_S L_S(f_{\theta},\tilde{D}^{\tilde{N}})
\end{equation}

In practice, when the classifier $f_\theta$ is trained iteratively, as is the case when using a neural network-based classifier, the above described applications of $P$, $T$, and $L_T$ are done batch-wise as outlined in Algorithm \ref{alg:strreggen}. 

The full class of data-based structural regularization methods are encompassed by this generic formulation, with particular methods varying in the choice of pseudo-labelling $P$, transformations $T_X$ and $T_Y$, and the selection of measure $R$. We next describe two important instances of structural regularization – consistency regularization and Mixup.

\begin{algorithm}[tb]
   \caption{Structural Regularization}
   \label{alg:strreggen}
\begin{algorithmic}
   \State {\bfseries Input:} A dataset $D^N=(X^N,Y^N)$, a family of models $\mathcal{F}=\{f_\theta:\theta\in\Theta\}$, a pseudo-labeling function $P$, a data transformation function $T=(T_\mathcal{X}, T_\mathcal{Y})$, a structural loss $L_S$, a structural loss weight $w_S$, and a primary task loss $L$ (e.g. MSE)
   \For{$t$ {\bfseries in} $\{1,...,T\}$}
   
   \State Draw a batch of size $m=m_L+m_U$:  $D^m_t=\{D^{m_L}_{L,t},D^{m_U}_{U,t}\}\subseteq D^N$
   
   \State Generate pseudo-labels:  $\hat{D}^m_t=\{D^{m_L}_{L,t},\hat{D}^{m_U}_{U,t}\}$, $\hat{D}^{m_U}_{U,t}=P(D^{m_U}_t,f_{\theta_{t-1}})$ and $f_{\theta_{t-1}}$ is an estimate of the classifier at $t-1$
   
   \State Generate a synthetic batch: $\tilde{D}^{\tilde{m}}_t=(\tilde{D}^{\tilde{m}_L}_{L,t},\tilde{D}^{\tilde{m}_U}_{U,t})=T(\hat{D}^m_t)$
   
   \State Evaluate loss: $L_T(f_{\theta_{t-1}},\tilde{D}^{\tilde{m}}_t)=L(f_{\theta_{t-1}},\tilde{D}^{\tilde{m}_L}_{L,t})+w_S L_S(f_{\theta_{t-1}},\tilde{D}^{\tilde{m}}_t)$
   \State Update model weights (e.g. gradient descent): $\theta_{t}=\theta_{t-1}-\alpha \nabla_{\theta} L_T(f_{\theta_{t-1}},\tilde{D}^{\tilde{m}}_t)$
   \EndFor
   \State {\bfseries Output:} $f_{\theta_T}$
\end{algorithmic}
\end{algorithm}

\subsection{Consistency Regularization}

Consistency regularization is a well-studied case of structural regularization and is used to introduce a class invariance structure with respect to specified transformations into model training. In this case, $T_\mathcal{X}: \mathcal{X}^N\to \mathcal{X}^N$, with $T_\mathcal{X} (X^N)=[vec(t_\mathcal{X} )](X^N)$, where $t_\mathcal{X}:\mathcal{X}^1\to \mathcal{X}^1$ is vectorized (denoted $vec(t_\mathcal{X})$) and applied individually to all $x\in X^N$, thus transforming each original observation’s features $x$ into synthetic features $\tilde{x}$.

Common transformations borrowed from supervised image classification include additive noise, random shifts, and random flips along with stronger transformations such as color and contrast adjustment \citep{77_Shorten2019ASO}. More recent work has also developed methods which learn a policy for combining several such transformations \citep{32_Cubuk2019AutoAugmentLA} \citep{33_Cubuk2020RandaugmentPA} \citep{75_UDAxie2020unsupervisedUDA}. These types of transformations can all be characterized by their reliance on prior domain knowledge (e.g. the knowledge that a shifted image of a dog remains a realistic sample from $p(x)$). On the other hand, domain agnostic transformations of the feature space limit the likelihood of producing unwanted synthetic observations by restricting themselves to taking only small departures from $x$ to generate $\tilde{x}$ \citep{111_Miyato2019VirtualAT} \citep{1_zhang2018mixup}. 

In contrast to transformations made directly to the observed feature space, another approach applies noise to learned representations of the observations, either using stochastic network architectures for the classifier $f_\theta$ \citep{105_dropout} \citep{106_Huangstochdepth}, or by perturbing observations in a learned latent space before mapping them back to the observed feature space \citep{7_mangla2020varmixup} \citep{16_8545506} \citep{8_yaguchi2019mixfeat}. 

To introduce invariance to feature transformation $T_\mathcal{X}$, the response transformation $T_\mathcal{Y}:\mathcal{Y}^N\to \mathcal{Y}^N$ is constructed as follows. $T_\mathcal{Y} (Y^N)=[vec(t_\mathcal{Y} )](Y^N)$, where
\begin{equation}\label{eq:3}
    \tilde{y}=t_{\mathcal{Y}}(y)=\begin{cases} 
    y,y\ne \emptyset\\
    \hat{y},y=\emptyset 
   \end{cases}.
\end{equation}
In the labelled case, taking the conditional class distribution target $\tilde{y}$ to be equal to the true label of the original observation $y$, the structural loss can be interpreted as preserving class-consistency between original and synthetic observation pairs $(x,\tilde{x})$ by assigning them both the label $y$. 

In the unlabeled case, an estimated class distribution $\hat{y}\in \hat{Y}^{n_U}_U$ is first generated for each original unlabeled observation $x\in X^{n_U}_U$ by the pseudo-labelling function $P(D^{n_U}_U,f_{\theta})=\hat{D}^{n_U}_U=(X^{n_U}_U,\hat{Y}^{n_U}_U)$.
Requiring consistency between $f_\theta (\tilde{x})$ and $\hat{y}$ results in a classifier $f_\theta$ that produces consistent predictions for $(x,\tilde{x})$ pairs without any knowledge about their true label. The form of the pseudo-labelling function which generates the estimate $\hat{y}$ varies by method. Choices for generating $\hat{y}$ have included the current best prediction of the classifier ($\hat{y}_t=f_{\theta_t} (x)$) \citep{115_PiLaine2017TemporalEF} \citep{111_Miyato2019VirtualAT}, an exponential moving average (EMA) of such predictions ($\hat{y}_t=\kappa \hat{y}_{t-1}+(1-\kappa)f_{\theta_t}(x)$) \citep{115_PiLaine2017TemporalEF}, or most recently an exponential moving average (EMA) of model weights across training ($\hat{y}_t = f_{\bar{\theta}_t}(x)$ where $\bar{\theta}_t = \kappa \bar{\theta}_{t-1}+(1-\kappa)\theta_t$) \citep{113_meanteach}. 

The label-preserving nature of consistency regularization connects it closely to the low-density separation and manifold assumptions which underpin many semi-supervised methods \citep{95_chapelle2006semi-supervised}. If one considers the feature transformation $T_\mathcal{X}$ as generating synthetic features local to those of an original observation as measured on the data manifold, then repeated application of $T_\mathcal{X}$ generates a high density region around each original observation. The response transformation $T_\mathcal{Y}$ then ensures that all observations in this region share a label. Thus the decision boundary is discouraged from passing through this high density region and is pushed away from the existing observations into a lower-density region.

The synthetic dataset $\tilde{X}^{\tilde{N}}$ generated using the pseudo-labelled dataset $\hat{D}^N$ and transformation function $T=(T_\mathcal{X},T_\mathcal{Y})$ is finally paired with a structural loss function (Eq. \ref{eq:2}) which operates on the transformed dataset and encourages the learning of a classifier that captures the invariances encoded in $\tilde{X}^{\tilde{N}}$. Common choices of the dissimilarity measures that define $L_S$ include mean squared error \citep{113_meanteach}, KL-divergence \citep{111_Miyato2019VirtualAT}, and cross-entropy loss \citep{1_zhang2018mixup} \citep{75_UDAxie2020unsupervisedUDA}. 

Consistency regularization training follows the general structural regularization training process (Algorithm \ref{alg:strreggen}) in this case using a transformation function that introduces consistency into the synthetic dataset $\tilde{D}^{\tilde{N}}$.

\subsection{Mixup}

Mixup regularization \citep{1_zhang2018mixup}, a special case of \citep{2_Chapelle2000VicinalRM} and related to \citep{35_Chawla2002SMOTESM} and  \citep{87_inoue2018data}, is a structural regularization method that applies a linear interpolation structure to the feature space between observations. In this case, $T_\mathcal{X}: \mathcal{X}^N\to \mathcal{X}^N, T_\mathcal{X} (X^N)=[vec(t_\mathcal{X} )](X^N,permute(X^N))$, where $t_\mathcal{X}:(\mathcal{X}^1,\mathcal{X}^1)\to \mathcal{X}^1$ is vectorized (denoted $vec(t_\mathcal{X} )$) and applied to randomly paired observations, thus transforming feature pairs of original observations $(x_i,x_j)$ into single synthetic features $\tilde{x}_{ij}$. In particular, the feature transformation produces a convex combination of the inputs as the synthetic features $\tilde{x}_{ij}$, with the mixing parameter $\lambda$ drawn from a Beta distribution parameterized by hyperparameter $\beta$ (Eq. \ref{eq:4}).
\begin{equation}\label{eq:4}
    \begin{split}
        \tilde{x}_{ij} = t_{\mathcal{X}}(x_i,x_j)=\lambda x_i +(1-\lambda)x_j \\
        \lambda \sim Beta(\beta,\beta)~~~~~~~~~~
    \end{split}
\end{equation}
Thus $T_\mathcal{X}$ yields a set of observation features $\tilde{X}^N$ that are interpolations between the original features $X^N$. To achieve a linear 
interpolation of the response, the response transformation $T_\mathcal{Y}: \mathcal{Y}^N\to \mathcal{Y}^N, T_\mathcal{Y} (Y^N)=[vec(t_\mathcal{Y} )](Y^N,permute(Y^N))$, with $t_\mathcal{Y}:(\mathcal{Y}^1,\mathcal{Y}^1)\to \mathcal{Y}^1$, applies the same mixing to the corresponding responses $y_i$, $y_j$ (Eq. \ref{eq:5}).
\begin{equation}\label{eq:5}
    \tilde{y}_{ij} = t_{\mathcal{Y}}(y_i,y_j)=\lambda y_i +(1-\lambda)y_j
\end{equation}

As in the consistency regularization case, extension to the semi-supervised setting is straightforward (Eq. \ref{eq:p_mu}). First, an estimated class distribution $\hat{y}\in \hat{Y}^{n_U}_U$ is generated for each original unlabeled observation $x\in X^{n_U}_U$ by the pseudo-labelling function $P(D^{n_U}_U,f_{\theta})=\hat{D}^{n_U}_U=(X^{n_U}_U,\hat{Y}^{n_U}_U)$. Then, this pseudo-label $\hat{y}$ is substituted for the response $y$ when it is unavailable. 
\begin{equation}\label{eq:p_mu}
    \begin{gathered}
    \tilde{y}_{ij} = t_{\mathcal{Y}}(y_i,y_j)=\lambda p(y_i) +(1-\lambda)p(y_j) \\
    p(y)=\begin{cases} 
        y,y\ne \emptyset \\
        \hat{y},y=\emptyset
        \end{cases}
    \end{gathered}
\end{equation}
Recent semi-supervised Mixup methods have defined the pseudo-label $\hat{y}$ as either $f_{\bar{\theta}_t}(x)$ where $\bar{\theta}_t = \kappa \bar{\theta}_{t-1} + (1-\kappa)\theta_t$ (as in ICT \citep{112_Verma2019ICT}) or have used more involved procedures (as in MixMatch \citep{80_mixmatch}). In both works, these Mixup transformation-based methods have favored using mean squared error as the structural loss. 

As with consistency regularization, model training using Mixup follows the general structural regularization training process (Algorithm \ref{alg:strreggen}) but using a transformation function that introduces this interpolative structure into the synthetic dataset $\tilde{D}^{\tilde{N}}$.

Connecting this interpolation transformation to the semi-supervised low-density separation and manifold assumptions \citep{95_chapelle2006semi-supervised}, we note the following. In the case that a pair of observations share a label, Mixup produces a similar effect to consistency regularization. With repeated application of $T$, and thus repeated draws from $Beta(\beta,\beta)$, the inter-observational space is made dense and the label of the endpoints is propagated along it. As a result the decision boundary is discouraged from separating the originally paired points. If instead the pair of observations have differing labels, the decision boundary is encouraged to lie exactly at the midpoint between the observations - far from the higher density endpoints and instead in the usually low density region between them.

\section{Epsilon Consistent Mixup}
In this section we present our proposed structural regularization method, Epsilon Consistent Mixup ($\epsilon$mu), and demonstrate experimentally its improvement in classification accuracy over competing semi-supervised methods. We also demonstrate $\epsilon$mu's  improvement of pseudo-label accuracy and predictive confidence over Mixup-based methods specifically. 

\subsection{Methodology}

Epsilon Consistent Mixup Regularization ($\epsilon$mu) is a structural regularization method which introduces an adaptive structure to the feature space between observations. In particular, $\epsilon$mu allows a combined structure of interpolation and consistency to be learned between observations in the feature space during training. As with Mixup, $\epsilon$mu generates synthetic features $\tilde{x}_{ij}$ by transforming pairs of original observations $(x_i,x_j)$ using the transformation $T_{\mathcal{X}}: \mathcal{X}^N\to \mathcal{X}^N, T_\mathcal{X} (X^N)=[vec(t_\mathcal{X})](X^N,permute(X^N))$, where $t_\mathcal{X}:(\mathcal{X}^1,\mathcal{X}^1)\to \mathcal{X}^1$ is vectorized (denoted $vec(t_\mathcal{X})$) and applied to randomly paired observations, and $t_\mathcal{X}$ is defined as in Eq. \ref{eq:4} above.

To produce the adaptive consistency-interpolation trade-off, the $\epsilon$mu response transformation is decoupled from the synthetic feature generation (using $T_\mathcal{X}$) and is defined as $T_\mathcal{Y}: \mathcal{Y}^N\to\mathcal{Y}^N, T_\mathcal{Y} (Y^N)=[vec(t_\mathcal{Y})](Y^N,permute(Y^N))$, with $t_\mathcal{Y}:(\mathcal{Y}^1,\mathcal{Y}^1)\to\mathcal{Y}^1$ and $t_\mathcal{Y}$ defined in Eq. \ref{eq:8} below. 
\begin{equation}\label{eq:8}
    \begin{gathered}
        t_{\mathcal{Y}}(y_i,y_j)=\eta_\epsilon(\lambda)p(y_i)+(1-\eta_{\epsilon}(\lambda))p(y_j)=\tilde{y}_{ij} \\
        \eta_\epsilon(\lambda)=\begin{cases}
        0,\lambda \leq \nu_\epsilon \\
        \frac{\lambda-\nu_\epsilon}{1-2\nu_\epsilon},\nu_\epsilon<\lambda<1-\nu_\epsilon \\
        1,\lambda \geq 1-\nu_\epsilon
        \end{cases} \\
        p(y)=\begin{cases} 
        y,y\ne \emptyset \\
        \hat{y},y=\emptyset
        \end{cases} 
        \nu_\epsilon = \frac{\epsilon}{\|x_i-x_j\|_2}
    \end{gathered}
\end{equation}

As with Mixup, the semi-supervised setting is accommodated using a pseudo-labelling function $P(D^{n_U}_U,f_{\theta})=\hat{D}^{n_U}_U=(X^{n_U}_U,\hat{Y}^{n_U}_U)$ which is used to generate an estimated class distribution $\hat{y}\in \hat{Y}^{n_U}_U$ for each original unlabeled observation $x\in X^{n_U}_U$. We follow \cite{112_Verma2019ICT} and generate $\hat{y}$ = $f_{\bar{\theta}_t}(x)$ where $\bar{\theta}_t = \kappa \bar{\theta}_{t-1} + (1-\kappa)\theta_t$.

\begin{figure}[tb]
    \centering
    \includegraphics[scale=0.3]{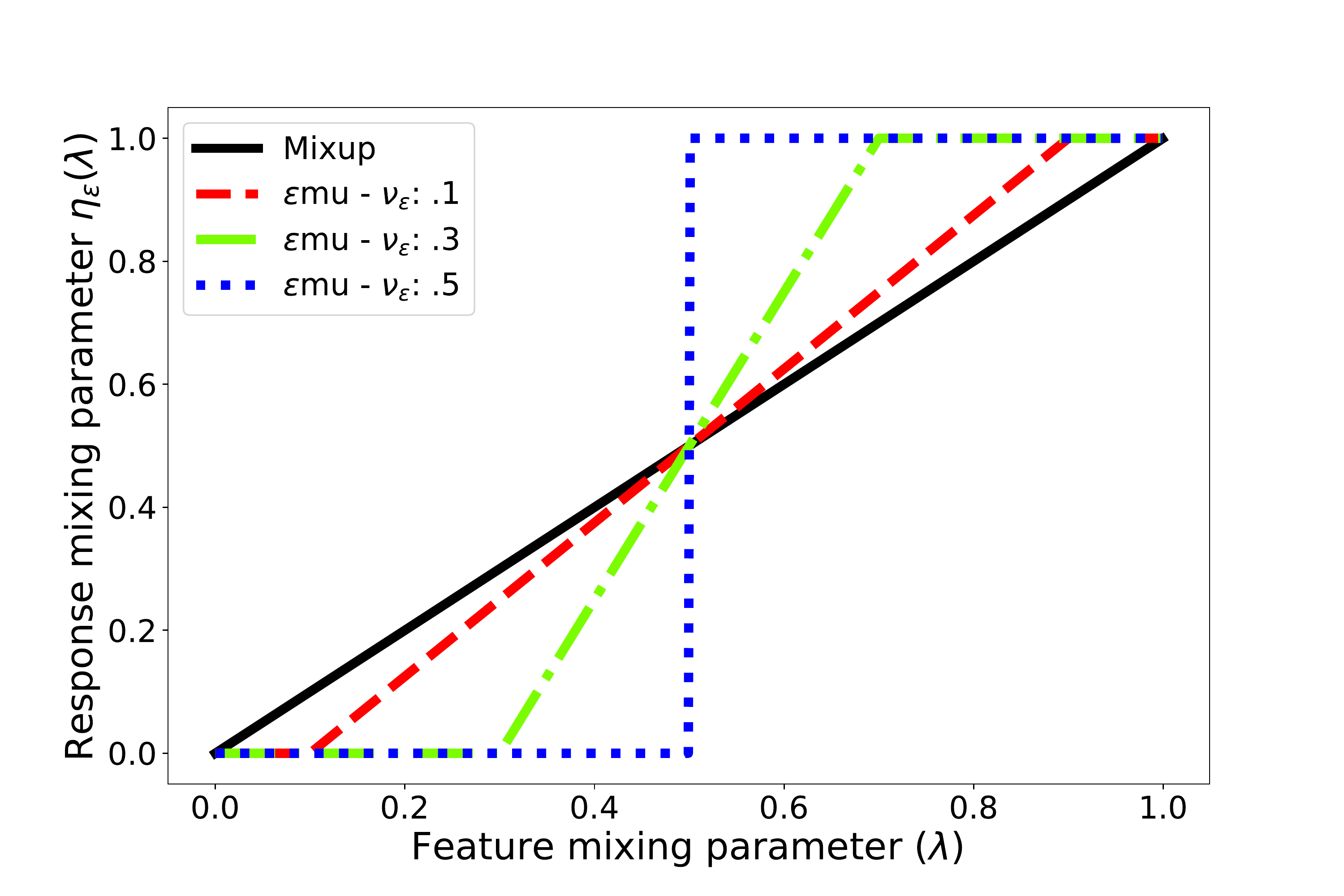}
    \caption{This figure illustrates how the $\epsilon$mu response mixing parameter $\eta_{\epsilon}(\lambda)$ relates to the feature mixing parameter $\lambda$ across several values for the rescaled consistency neighborhood radius $\nu_\epsilon$.}
    \label{fig:lambdaveta}
\end{figure}

As compared to pure Mixup, $\epsilon$mu decouples the mixing of the features $(x_i,x_j)$ from the mixing of the responses $(y_i,y_j)$. In particular, response interpolation is prevented within the $\epsilon$-neighborhood in the feature space around each observation via the rescaled consistency neighborhood radius $\nu_\epsilon$. Fig. \ref{fig:lambdaveta} demonstrates how the mixing of the responses is affected by changes in $\lambda$ for differing $\nu_\epsilon$. Outside this $\nu_\epsilon$ neighborhood, interpolation proceeds linearly as in Mixup. Within the $\epsilon$-neighborhood of each observation however, consistency with the observed response $y$ is maintained. The intuition being that synthetic images indistinguishable from the originals should remain in the same class. This local consistency also has an entropy reducing effect on the predictions as demonstrated in the experiments section. 

The trade-off dynamic between interpolation and consistency enables the parameter $\epsilon$, which governs the size of consistency neighborhoods, to be learned during training. As a consequence, $\epsilon$mu has the flexibility to revert to traditional Mixup (with $\nu_\epsilon=0$) or to generate only consistency in the direction of existing observations ($\nu_\epsilon \geq 0.5$). The adaptive nature of $\epsilon$ has the further advantage of presenting no additional tuning burden compared with Mixup.

Although any dissimilarity measure can be used to define the structural loss operating on the $\epsilon$mu transformed data, we follow \citep{113_meanteach} and \citep{80_mixmatch} in defining $L_S$ as the mean squared error. 

\begin{figure}[tb]
    \centering
    \includegraphics[scale=0.3]{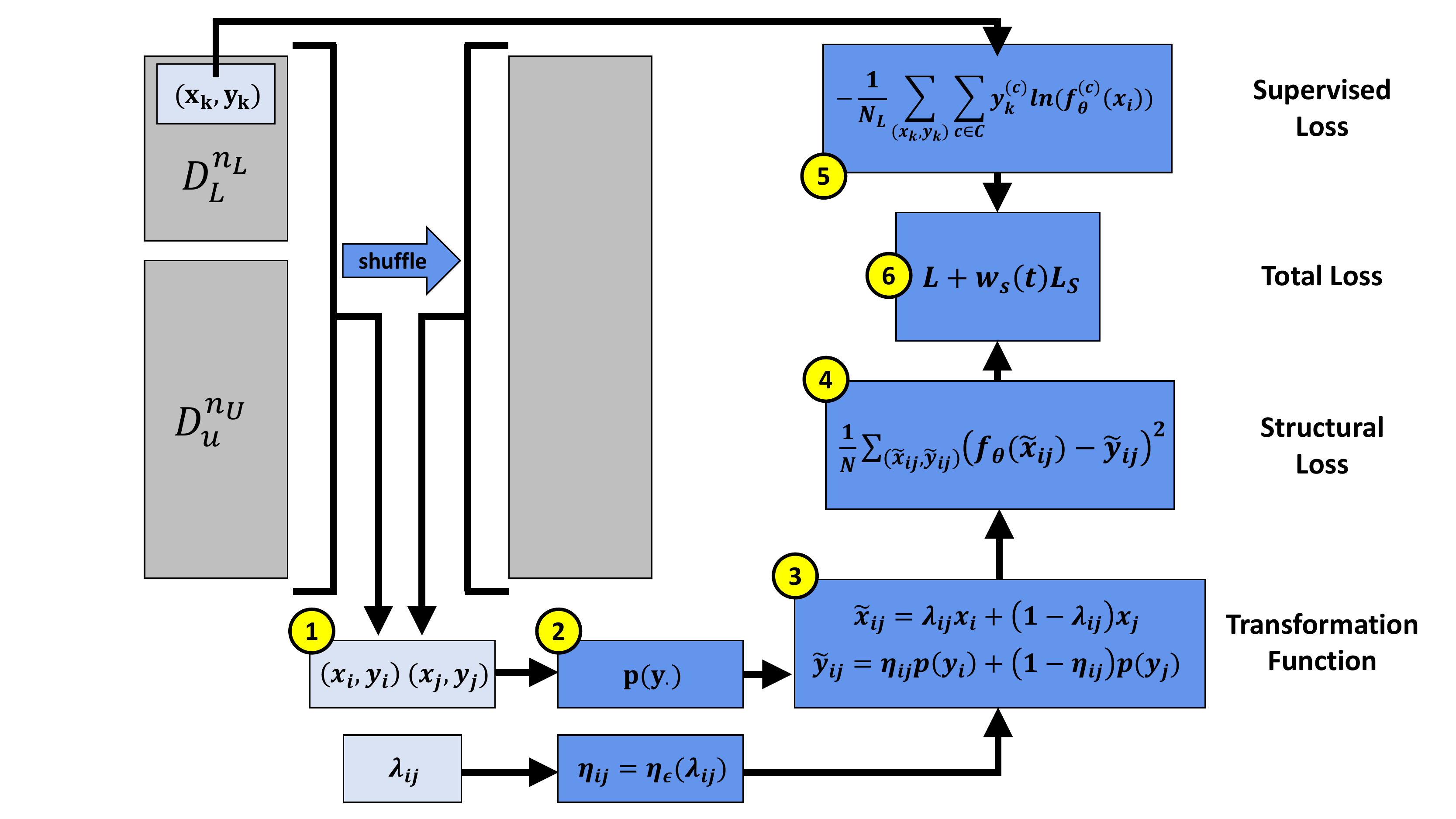}
    \caption{This figure summaries the $\epsilon$mu procedure. In Step 1, data is permuted and observations are randomly paired and assigned a randomly drawn feature interpolation parameter $\lambda$. In Step 2, psuedo-labels are generated and the response interpolation parameter $\eta$ is computed. In Step 3, the synthetic data are generated. In Step 4, the structural loss, MSE in this case, is calculated using the synthetic data. In Step 5, the original labeled observations are used to calculate the supervised loss, cross-entropy in this case. Finally, in Step 6, the two losses are combined in the total loss.}
    \label{fig:emualg}
\end{figure}

Model training using $\epsilon$mu, summarized in Fig. \ref{fig:emualg}, follows the general structural regularization training process (Algorithm \ref{alg:strreggen}) in this case using the transformation function defined in Eq. \ref{eq:8} to produce $\tilde{D}^{\tilde{N}}$.

\subsection{Experimental Results}
We first demonstrate the effectiveness of the proposed $\epsilon$mu regularization in the semi-supervised setting using two standard benchmark datasets, CIFAR10 \citep{116_cifar10Krizhevsky2009LearningML} and SVHN \citep{117_svhnNetzer2011ReadingDI}. We compare $\epsilon$mu classification accuracy to several types of semi-supervised method: the consistency regularization based methods $\Pi$-Model \citep{115_PiLaine2017TemporalEF}, Mean Teacher \citep{113_meanteach}, and Virtual Adversarial Training (VAT) \citep{111_Miyato2019VirtualAT}, the solely interpolation-based Interpolation Consistency Training (ICT) method \citep{112_Verma2019ICT}), and the agglomerative interpolation-based MixMatch method \citep{80_mixmatch}. The comparison of $\epsilon$mu with ICT is particularly important as the two methods use only interpolation to achieve semi-supervised training and differ only in ICT's use of Mixup instead of $\epsilon$mu interpolation. This similarity thus allows comparison between ICT and $\epsilon$mu performance to be isolated and attributable solely to $\epsilon$mu's consistency-interpolation tradeoff mechanism. Similarly, the relative effect of $\epsilon$mu to Mixup in a more complex multi-method approach such as MixMatch is achieved by comparing MixMatch (using Mixup) to MixMatch using $\epsilon$mu-based interpolation as a plug-in replacement to Mixup (MixMatch + $\epsilon$mu).

We then describe two further sets of experiments which highlight two sources of $\epsilon$mu's superior classification performance. In the first, we measure the distance between estimated true response values for interpolated observations and those generated by $\epsilon$mu and Mixup. These experiments demonstrate that $\epsilon$mu produces higher quality interpolated response targets ($\tilde{y}$) than Mixup while also simultaneously establishing the interpolative structure on a larger region of the feature space than Mixup. In the second set of experiments, we show that the local consistency of $\epsilon$mu has a strong entropy minimizing effect on classifier predictions as compared with Mixup, an effect known to produce better classification performance \citep{22_entrmin} and often targeted explicitly \citep{80_mixmatch} \citep{75_UDAxie2020unsupervisedUDA}. 
\paragraph{Datasets}
As is standard practice \citep{118_realistOliver2018RealisticEO}, the semi-supervised settings used in the experiments are created starting with the fully labelled CIFAR10 and SVHN datasets and ``masking" all but a small number of the labels contained in the training data. The three levels of semi-supervised labelling that were tested are 40 total labels (4 per class), 250 total labels (25 per class), and 500 total labels (50 per class).

The CIFAR10 dataset consists of 60K 32x32 color images of ten classes (airplanes, frogs, etc.). The data is split into a training set of 50K images and test set of 10K images. In our experiments, we further split the training dataset into a 49K dataset which is partially unlabeled and used for training and a fully labeled 1K validation set which is used to select hyperparameters.

The SVHN dataset consists of 32x32 color images of ten classes (the digits 0-9) split into a training set of 73,257 images and a test set of 26,032 images\footnote{Although the SVHN dataset makes available an additional 531,131 images, these are not used in our experiments}. In our experiments, the training dataset is further split into a 72,257 image dataset which is partially unlabeled and used for training, and a fully labeled 1K validation dataset used for hyperparameter selection. 

For both datasets, the image features are rescaled to the interval [-1,1]. Standard weak augmentation is also applied as in \citep{80_mixmatch}. The CIFAR10 images are horizontally flipped with probability 0.5, reflection padded by 2 pixels in the either direction and then randomly cropped to the original 32x32 dimensions. The SVHN images are reflection padded by 2 pixels in the either direction and then randomly cropped to the original 32x32 dimensions. 
\paragraph{Implementation Details}
All experiments are conducted using the 1.47M parameter “Wide ResNet-28” model proposed in \citep{119_Zagoruyko2016WideRN}. We follow the training procedure outlined in \citep{80_mixmatch}. In particular, a constant learning rate of $0.02$ is used throughout training and prediction is done using the exponential moving average of the model weights with decay rate 0.999. Training differs for ICT and $\epsilon$mu from \citep{80_mixmatch} in that we use only 1K observations for hyperparameter tuning and the error rate of the final epoch is reported, as opposed to the median rate of several checkpoints. Unless otherwise specified, models are trained for 409,600 batches with each batch comprised of 64 labelled and 64 unlabeled observations. One epoch is defined as 1024 batches. The loss weight $w_S (t)$ for the structural loss is increased linearly for the first 16 epochs to a maximal weight $w_S$. 
\paragraph{Baseline Models}
We provide comparisons to several consistency based semi-supervised methods, $\Pi$-model \citep{115_PiLaine2017TemporalEF}, Mean Teacher \citep{113_meanteach}, and VAT \citep{111_Miyato2019VirtualAT}, as implemented in \citep{80_mixmatch}. We also compare $\epsilon$mu performance to Interpolation Consistency Training (ICT) \citep{112_Verma2019ICT} which offers a direct comparison between the effects of Mixup and $\epsilon$mu. ICT is re-implemented using the optimization methods and model described above to provide a fair comparison to $\epsilon$mu. This re-implementation with the new training scheme provides ICT a significant performance gain compared to that reported in \citep{112_Verma2019ICT}. Finally, MixMatch \citep{80_mixmatch}, a more holistic method which incorporates multiple semi-supervised training techniques, is compared against MixMatch + $\epsilon$mu, the corresponding method replacing Mixup with $\epsilon$mu interpolation. 
\paragraph{Hyperparameters}
$\epsilon$mu regularization requires tuning of several hyperparameters, namely the maximal structural loss weight $w_S$, the Beta mixing distribution parameter $\beta$, the initialization value for the neighborhood radius $\epsilon$, and the L2 regularization penalty weight (referred to as the weight decay). Matching the search space from \citep{112_Verma2019ICT}, a grid search is done on the following parameter combinations: $\{(\beta,w_s): \beta\in\{0.1,0.2,0.5,1.0\}$ and $w_S\in\{1,10,20,100\}\}$. Training is very robust to the initialization value for $\epsilon$ with the only restriction being it fall within the range of inter-image distances observed in the sample. In practice $\epsilon$ is taken to be 10 for both CIFAR10 and SVHN, corresponding to an estimated 27.8\% and 36.0\% respectively of the distance between random pairs of observations on average. As noted in \citep{80_mixmatch}, training is found to be sensitive to the weight decay. Thus, for each $(\beta,w_S)$ pair, a range of values is searched; $\{$1.2e-4, 2.4e-4$\}$ for CIFAR10 and $\{$3.6e-4, 4.0e-4, 6.0e-4, 1.0e-3, 1.2e-3, 1.8e-3$\}$ for SVHN. Optimal hyperparameter values are listed in Tables \ref{table:appendixmodelsettings_cifar10ictemu} and \ref{table:appendixmodelsettings_svhnictemu}.

Due to the high number of MixMatch hyperparameters, MixMatch + $\epsilon$mu is tested using the best reported settings from \citep{80_mixmatch}, with the exception of the Beta mixing distribution parameter $\beta$. In the case of $\beta$, a value of $1.0$ is used for both datasets motivated by the observation from ICT vs. $\epsilon$mu experiments that $\epsilon$mu consistently prefers a more uniform mixing distribution (i.e. a larger $\beta$ value). Optimal hyperparameter values are listed in Tables \ref{table:appendixmixmatchmodelsettings_cifar10} and \ref{table:appendixmixmatchmodelsettings_svhn}.

\paragraph{Classification Performance}
Classification performance was assessed in three semi-supervised settings, one with 40 total labelled observations, another with 250, and lastly one with 500 (i.e. $n_L\in\{40,250,500\}$). For each level of labelling, several different $(D^{n_L}_L,D^{n_U}_U)$ splits were randomly generated. Models were trained on each split and the average performance and standard deviation of the splits was reported. For the re-implemented ICT and $\epsilon$mu results, performance was measured using three splits with the best hyperparameter settings reported in Tables \ref{table:appendixmodelsettings_cifar10ictemu} and \ref{table:appendixmodelsettings_svhnictemu}. MixMatch and MixMatch + $\epsilon$mu results were reported using 5 splits as in \citep{80_mixmatch} \citep{46_fxmatch}. Ten splits were used for $\Pi$-Model, Mean Teacher, and VAT performance assessment as reported in \citep{75_UDAxie2020unsupervisedUDA}.

On the CIFAR10 data, $\epsilon$mu outperformed the $\Pi$-model, Mean Teacher, and VAT methods by significant margins in all label settings (Table \ref{table:1}). $\epsilon$mu also significantly outperformed ICT in all settings. As ICT and $\epsilon$mu differ methodologically only in ICT’s use of Mixup as opposed to $\epsilon$mu for synthetic response generation, this ablative comparison allows attribution of all relative performance gains to $\epsilon$mu's unique consistency-interpolation tradeoff, showing that $\epsilon$mu is indeed responsible for significant accuracy improvement. Comparison was also made between MixMatch, an agglomerative method which utilizes additional semi-supervised techniques in addition to structural regularization, and MixMatch + $\epsilon$mu, the corresponding $\epsilon$mu-based method. In this comparison the two methods performed comparably, with MixMatch + $\epsilon$mu showing marginal outperformance of MixMatch in low-label settings. However, performance differences between the two were not significant - likely due to the higher variance of the less stable MixMatch-based approaches.

On the CIFAR10 dataset, the $\epsilon$mu learned consistency neighborhood $\epsilon$ was on average 7.45, 8.80, and 8.71 in the 40, 250, and 500 label settings respectively, which corresponded to approximately 20.6\%, 24.3\%, and 24.1\% of the distance between two random pairs of CIFAR10 images on average (Table \ref{table:appendixmodelsettings_cifar10ictemu}). Interestingly, the learned neighborhoods for MixMatch + $\epsilon$mu were directionally consistent but much larger at approx. 12.67 (35.1\%) (Table \ref{table:appendixmixmatchmodelsettings_cifar10}) suggesting that the more complex method preferred stronger consistency regularization. These large learned consistency radii suggest that $\epsilon$mu's flexibility was indeed advantageous in more faithfully modeling the response structure between observations.

On the SVHN dataset, $\epsilon$mu again significantly outperformed the $\Pi$-model and VAT methods across all settings (Table \ref{table:1}). It performed comparably with Mean Teacher in the 250 label setting, but was outperformed in the 500 label setting. Although it either surpassed (in the 250 label setting) or matched (in the 40 and 500 label settings) ICT accuracy on this dataset, $\epsilon$mu did not show as clear an advantage over ICT as it did on the more complex CIFAR10 data. As SVHN classification is a simpler task, we hypothesize that the added flexibility offered by $\epsilon$mu was not critical to fit the data well in this case, resulting in the similar performance between ICT and $\epsilon$mu. Evidence for this hypothesis is also reflected in the small average $\epsilon$ values of 4.32, 3.81, and 3.55 learned in the 40, 250, and 500 label settings respectively. These $\epsilon$ values corresponded to approximately 15.6\%, 13.7\%, and 12.8\% of the distance between two pairs of SVHN images on average (Table \ref{table:appendixmodelsettings_svhnictemu}). As with the CIFAR10 data, these learned consistency neighborhoods were directionally consistent but smaller than those learned by MixMatch + $\epsilon$mu at approx. 5.13 (18.5\%) (Table \ref{table:appendixmixmatchmodelsettings_svhn}). Suggesting again that the more complex MixMatch prefers stronger consistency regularization. On this dataset, $\epsilon$mu significantly outperformed both MixMatch and MixMatch + $\epsilon$mu in the lowest 40 label setting and remained competitive in the 250 and 500 label settings. 
We hypothesize that the relatively stronger performance of $\epsilon$mu and ICT as compared to the MixMatch-based methods in the 40 label setting is due to the difficulty MixMatch had learning in low label settings generally, coupled with the relatively simple SVHN classification task readily learned by $\epsilon$mu and ICT without the complexity of MixMatch.
As in the CIFAR10 experiments, MixMatch + $\epsilon$mu performed well as a plug-in replacement for a Mixup-based MixMatch. 

\begin{table*}[t]%
\caption{These tables contain the test error rates for stand-alone methods $\Pi$-Model, Mean Teacher, VAT, ICT, and $\epsilon$mu, and agglomerative methods MixMatch and MixMatch + $\epsilon$mu. The left table contains CIFAR10 error rates for the 40, 250, and 500 total label settings. Similarly, the right table contains SVHN error rates for the same settings.}
\centering
\normalsize
\begin{tabular}{ |p{2.1cm}||p{1.6cm}|p{1.6cm}|p{1.6cm}|  }
 \hline
 \multicolumn{4}{|c|}{CIFAR10 Misclassification Error Rates} \\
 \hline
 Method & 40 Labels & 250 Labels & 500 Labels\\
 \hline
 $\Pi$-Model& -& $53.02\pm 2.05$& $41.82\pm 1.52$\\
 Mean Teacher& -& $47.32\pm 4.71$& $42.01\pm 5.86$\\ 
 VAT& -& $36.03\pm 2.82$& $26.11\pm 1.52$\\ 
 \hline
 ICT& $47.77\pm 2.23$& $16.76\pm 0.51$& $11.24\pm 0.18$\\ 
 $\epsilon$mu& $37.17\pm1.87$& $12.56\pm 1.42$& $10.19\pm 0.12$\\ 
 \hline
 MixMatch& $47.54\pm 11.50$& $11.08\pm 0.87$& $9.65\pm 0.94$\\ 
 MixMatch + $\epsilon$mu& $32.45\pm 10.36$& $10.43\pm 0.56$& $9.99\pm 0.96$\\ 
 \hline
\end{tabular}
\begin{tabular}{ |p{2.1cm}||p{1.6cm}|p{1.6cm}|p{1.6cm}|  }
 \hline
 \multicolumn{4}{|c|}{SVHN Misclassification Error Rates} \\
 \hline
 Method & 40 Labels & 250 Labels & 500 Labels\\
 \hline
 $\Pi$-Model& -& $17.65\pm 0.27$& $11.44\pm 0.39$\\ 
 Mean Teacher& -& $6.45\pm 2.43$& $3.82\pm 0.17$\\ 
 VAT& -& $8.41\pm 1.01$& $7.44\pm 0.79$\\ 
 \hline
 ICT& $25.52\pm 5.66$& $5.00\pm 0.28$& $4.18\pm 0.06$\\ 
 $\epsilon$mu& $24.24\pm 4.46$& $4.37\pm 0.19$& $4.10\pm 0.05$\\ 
 \hline
 MixMatch& $42.55\pm 14.53$& $3.78\pm 0.26$& $3.64\pm 0.46$\\ 
 MixMatch + $\epsilon$mu& $42.37\pm 17.43$& $4.40\pm 0.87$& $3.94\pm 0.34$\\ 
 \hline
\end{tabular}
\label{table:1}
\end{table*}

\paragraph{Synthetic Label Quality}

In order to better understand the ways in which $\epsilon$mu differs from Mixup, we conducted an experiment measuring the discrepancy between the synthetic labels generated by these methods and the estimated true labels $\hat{y}$ of each synthetic observation in the 250 label setting for the CIFAR10 dataset. This setting was selected because it offered a clean contrast between $\epsilon$mu and ICT - the methods performed differently in this case, and training was more stable relative to the lowest 40 label setting. The estimate $\hat{y}$ was used as a proxy for the true labels of the synthetic interpolated images as labels for them do not exist. The estimated true labels used in these experiments were the labels predicted by a fully supervised Wide ResNet-28 model trained on the full set of labelled data with weak image augmentation and a small weight decay of 0.0001. To compare $\hat{y}$ with the labels produced by $\epsilon$mu ($\tilde{y}_{\epsilon mu}$) and those produced by ICT using Mixup ($\tilde{y}_{mu}$), we use $Q_{\cdot} = \|\tilde{y}_{\cdot}-\hat{y}\|_2^{-1}$, the inverse of the L2 distance, which we refer to as the quality of the labeling.

\begin{figure}[tb]
\centering
\captionsetup[sub]{skip=0pt,font=large,labelformat=empty}
  \begin{subfigure}[t]{0.475\textwidth}
    \centering
    \includegraphics[width=\textwidth]{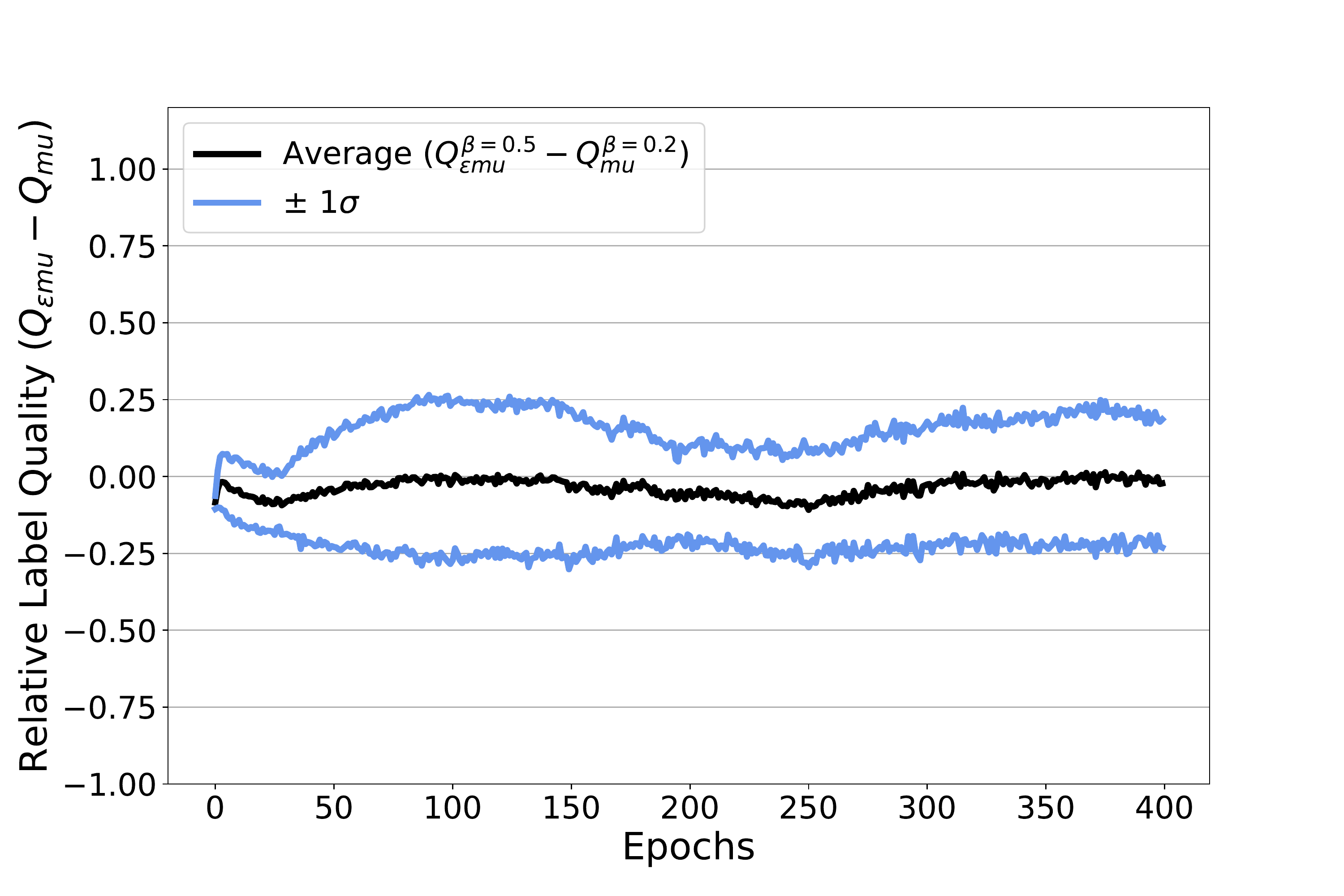}
    \caption{\textbf{(4a)}}
    \label{fig-a}
  \end{subfigure}\hfill
  \begin{subfigure}[t]{0.475\textwidth}
    \centering
    \includegraphics[width=\textwidth]{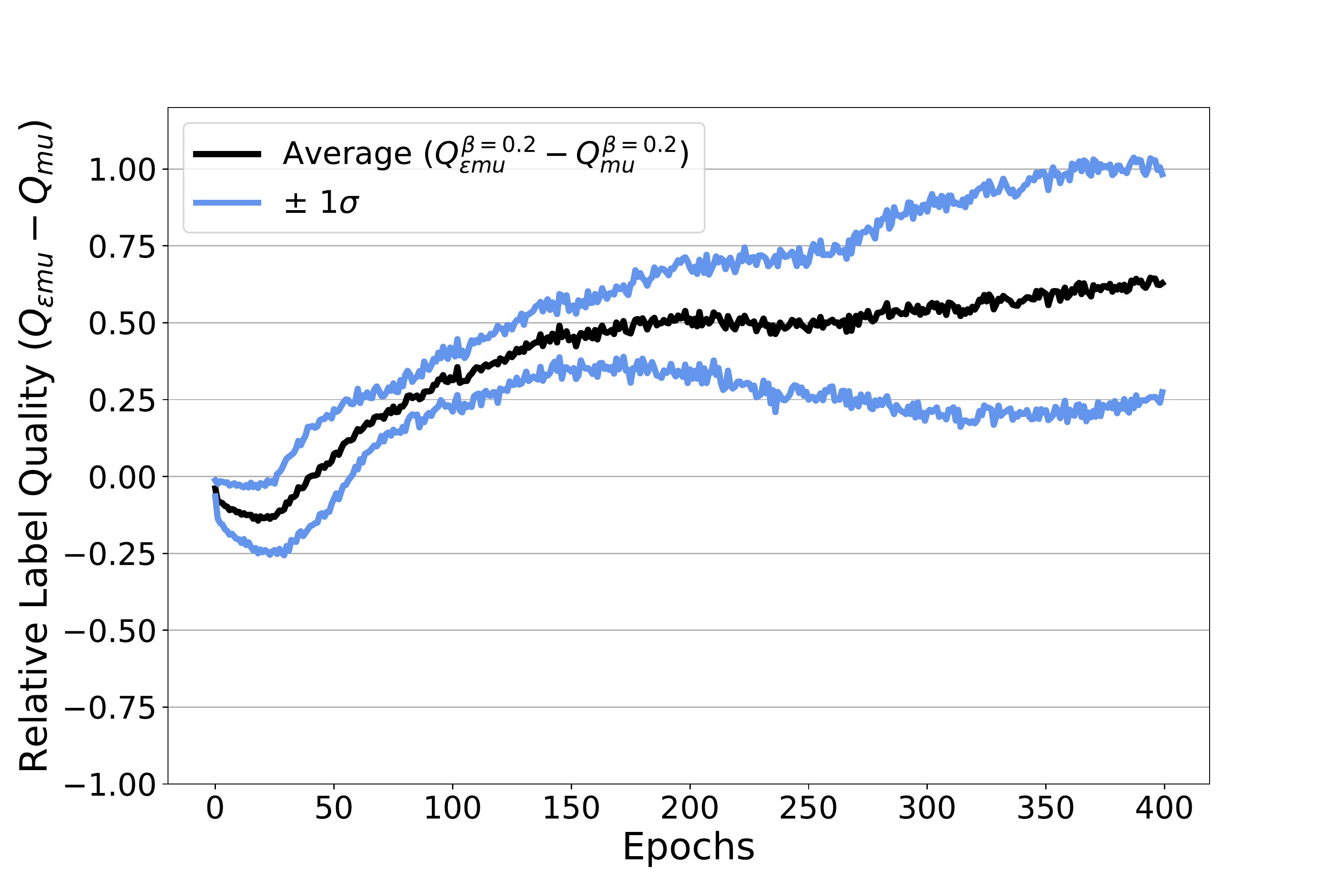}
    \caption{\textbf{(4b)}}
    \label{fig-b}
  \end{subfigure}
  \caption{The plot on the left (\textbf{4a}) shows that the label quality of the most accurate $\epsilon$mu model ($\beta=0.5$) is no different than that of the most accurate Mixup-based ICT model ($\beta=0.2$) in the 250 labelled CIFAR10 setting. The plot on the right (\textbf{4b}) shows that the label quality of the $\epsilon$mu model exceeds that of the highest performing ICT model when the mixing parameter most advantageous to ICT ($\beta=0.2$) is used for both models. The metric average and standard deviation bands are obtained by training models on 5 random $(D^{n_L}_L,D^{n_U}_U)$ splits.} 
  \label{fig:quality}
\end{figure}

The quality of the labelling was assessed as follows. First, the label qualities of both $\tilde{y}_{\epsilon mu}$ and $\tilde{y}_{mu}$ were measured across training and were found to improve as the classifiers themselves improved. Next, the relative quality of the two interpolation schemes, defined as $Q_{\epsilon mu}-Q_{mu}$, was measured in two ways. In the first, the label quality of the best $\epsilon$mu model was compared to the label quality of the best ICT model. This comparison revealed that the two best models for each method produced the same quality of labelling (Fig \ref{fig:quality}a). However, the best $\epsilon$mu model used a higher mixing distribution parameter ($\beta=0.5$) than the best ICT model ($\beta=0.2$), indicating that on average it produced interpolations farther from the original observations. Thus, while the quality of the labelling was equal, the $\epsilon$mu model established its interpolative structure on a larger region of the feature space. In the second comparison, the best ICT model was compared with the $\epsilon$mu model using this ICT model’s hyperparameter settings (including the same $\beta=0.2$). This comparison showed that when interpolation is restricted to comparable regions, $\epsilon$mu produces higher quality labels compared to the Mixup interpolation used in ICT (Fig \ref{fig:quality}b). These comparisons demonstrate two sources of $\epsilon$mu’s performance improvement, the generation of higher quality labeling and the ability to impose it on a larger part of the feature space.

\begin{figure}[tb]
    \centering
    \includegraphics[scale=0.3]{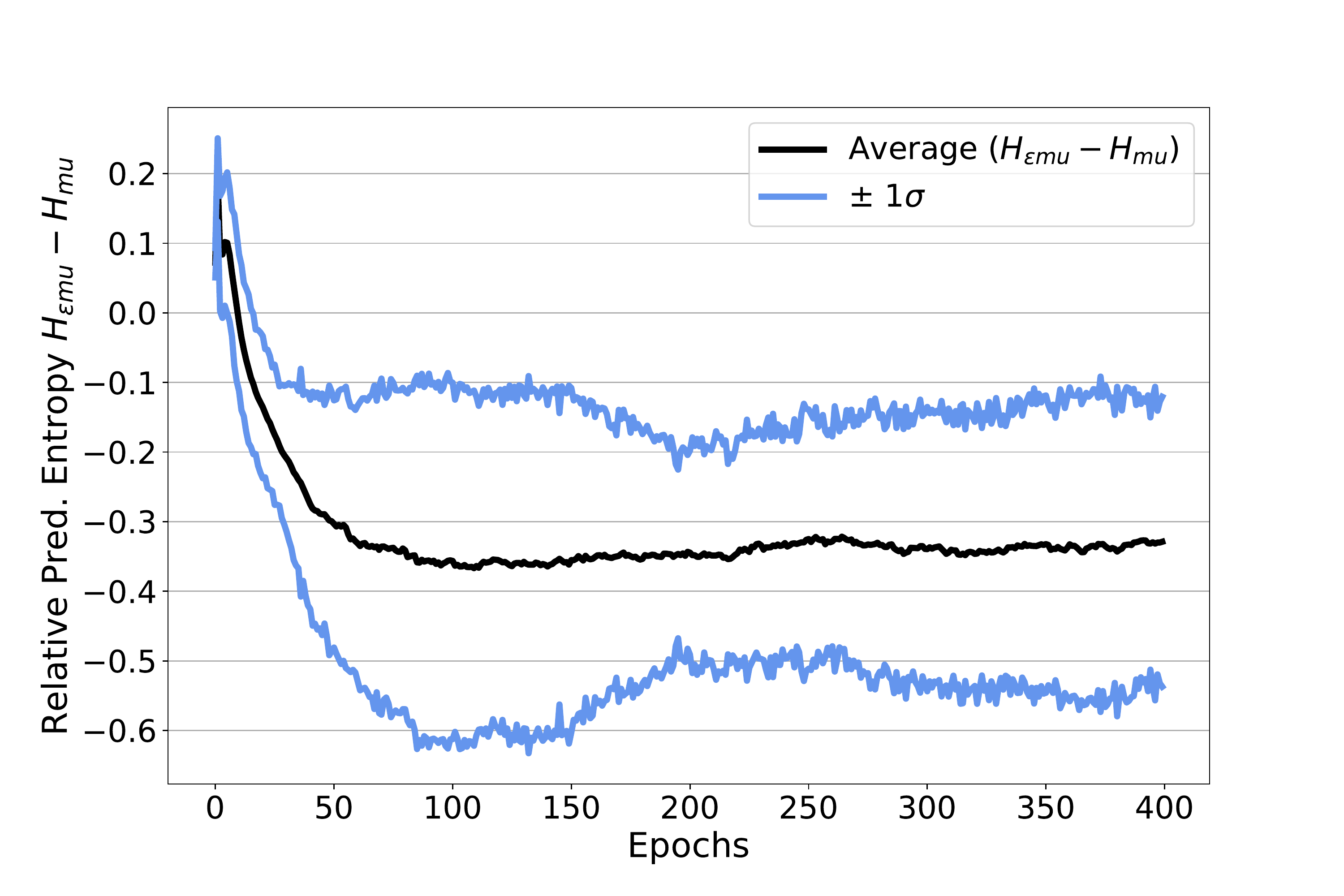}
    \caption{This figure shows the average difference between the predicted label distribution entropy of the most accurate model trained using $\epsilon$mu ($H_{\epsilon mu}$) and the corresponding entropy produced by the most accurate Mixup-trained model ($H_{mu}$) in the CIFAR10 250 label setting. The negative difference indicates that the $\epsilon$mu-trained model generates more confident predictions. The metric average and standard deviation bands are obtained by training models on 5 random $(D^{n_L}_L,D^{n_U}_U)$ splits.}
    \label{fig:entropy}
\end{figure}

\paragraph{Entropy Minimization}
In contrast to the synthetic label quality experiment which assessed differences in the $\epsilon$mu and Mixup training mechanisms, in this experiment we assess the difference between the resultant models trained by the two methods. Again in the 250 label setting using the CIFAR10 dataset, we measured the entropy of the predicted conditional class distributions ($f_{\theta}(x)$) of models trained using each method. We found that $\epsilon$mu-trained classifiers produced consistently and significantly higher confidence (and thus lower entropy) in predictions on the training data (Fig. \ref{fig:entropy}). This result suggests that the consistency neighborhoods learned using $\epsilon$mu produce a sharpening effect on model predictions similar to that generated by explicit entropy penalization methods. As these entropy minimization methods have been shown to be complementary to consistency and interpolation based structural regularization \citep{80_mixmatch} \citep{75_UDAxie2020unsupervisedUDA}, the gains in $\epsilon$mu’s performance over Mixup are likely due partly to this two-for-one effect. 

\section{Conclusion}

In this paper we introduced $\epsilon$mu, an adaptive structural regularization method that combines Mixup style interpolation with consistency preservation in the Mixup direction. We demonstrated $\epsilon$mu's improvement over Mixup in semi-supervised image classification and showed that $\epsilon$mu produces more confident predictions and more accurate synthetic labels over a larger region of the feature space. 

The effectiveness of $\epsilon$mu in this setting suggests several promising directions for future development. In the direction of methodology, further improvements to $\epsilon$mu's adaptive tradeoff mechanism could be targeted by, for example, learning an observation-specific consistency neighborhood generator, i.e. $\epsilon_i=f_{\theta}(x_i)$, or by relaxing the response-mixing function to a class of smooth parameterized interpolation functions, e.g. regularized beta functions. In an applied direction, adaptation of $\epsilon$mu to other domains such as text, genomics, and even tabular data appears promising, with the likely challenge being adjusting the feature-mixing procedure, perhaps using intermediate representations \citep{3_2019_manifoldmixup}.


\begin{table*}
\caption{This table summarizes the best model hyperparameter values for ICT and $\epsilon$mu on the CIFAR10 dataset}
\centering
\normalsize

\begin{tabular}{ |c|c||p{.5cm}|p{.5cm}|c|c|p{1.9cm}| }
 \hline
 \multicolumn{7}{|c|}{CIFAR10 Best Model Hyperparameter Settings} \\
 \hline
 \multirow{2}{*}[-.9em]{Method} & \multirow{2}{*}[-.9em]{Labels} & \multirow{2}{*}[-.9em]{$w_S$} & \multirow{2}{*}[-.9em]{$\beta$} & \multirow{2}{*}[-.9em]{Weight Decay} & \multicolumn{2}{|c|}{Avg. learned $\epsilon$}\\
 \cline{6-7}
 & & & & & Magnitude & \% of Avg. Inter-Image Distance\\
 \hline
 ICT & 40 & 20 & 0.2 & 1.2e-4 & 0.0 & -\\
 \hline
 $\epsilon$mu & 40 & 10 & 1.0 & 2.4e-4 & 7.45 & 20.6\%\\
 \hline
 ICT & 250 & 50 & 0.2 & 2.4e-4 & 0.0 & -\\
 \hline
 $\epsilon$mu & 250 & 100 & 0.5 & 1.2e-4 & 8.80 & 24.3\%\\
 \hline
 ICT & 500 & 100 & 0.2 & 1.2e-4 & 0.0 & -\\
 \hline
 $\epsilon$mu & 500 & 20 & 0.5 & 1.2e-4 & 8.71 & 24.1\%\\
 \hline
\end{tabular}
\label{table:appendixmodelsettings_cifar10ictemu}
\end{table*}

\begin{table*}
\caption{This table summarizes the best model hyperparameter values for ICT and $\epsilon$mu on the SVHN dataset}
\centering
\normalsize
\begin{tabular}{ |c|c||p{.5cm}|p{.5cm}|c|c|p{1.9cm}| }
 \hline
 \multicolumn{7}{|c|}{SVHN Best Model Hyperparameter Settings} \\
 \hline
 \multirow{2}{*}[-.9em]{Method} & \multirow{2}{*}[-.9em]{Labels} & \multirow{2}{*}[-.9em]{$w_S$} & \multirow{2}{*}[-.9em]{$\beta$} & \multirow{2}{*}[-.9em]{Weight Decay} & \multicolumn{2}{|c|}{Avg. learned $\epsilon$}\\
 \cline{6-7}
 & & & & & Magnitude & \% of Avg. Inter-Image Distance\\
 \hline
 ICT & 40 & 1 & 0.1 & 1.8e-3 & 0.0 & -\\
 \hline
 $\epsilon$mu & 40 & 1 & 1.0 & 1.0e-3 & 4.32 & 15.6\%\\
 \hline
 ICT & 250 & 20 & 0.1 & 6.0e-4 & 0.0 & -\\
 \hline
 $\epsilon$mu & 250 & 20 & 0.1 & 3.6e-4 & 3.81 & 13.7\%\\
 \hline
 ICT & 500 & 20 & 0.1 & 3.6e-4 & 0.0 & -\\
 \hline
 $\epsilon$mu & 500 & 50 & 0.1 & 3.6e-4 & 3.55 & 12.8\%\\
 \hline
\end{tabular}
\label{table:appendixmodelsettings_svhnictemu}
\end{table*}


\begin{table*}
\caption{This table summarizes the best model hyperparameter values for MixMatch and MixMatch + $\epsilon$mu on the CIFAR10 dataset}
\centering
\normalsize

\begin{tabular}{ |p{2.2cm}|c||p{.5cm}|p{.5cm}|c|c|p{1.9cm}| }
 \hline
 \multicolumn{7}{|c|}{CIFAR10 Best Model Hyperparameter Settings} \\
 \hline
 \multirow{2}{*}[-.9em]{Method} & \multirow{2}{*}[-.9em]{Labels} & \multirow{2}{*}[-.9em]{$w_S$} & \multirow{2}{*}[-.9em]{$\beta$} & \multirow{2}{*}[-.9em]{Weight Decay} & \multicolumn{2}{|c|}{Avg. learned $\epsilon$}\\
 \cline{6-7}
 & & & & & Magnitude & \% of Avg. Inter-Image Distance\\
 \hline
 MixMatch & 40 & 75 & 0.75 & 4e-5 & 0.0 & -\\
 \hline
 MixMatch + $\epsilon$mu & 40 & 75 & 1.0 & 4e-5 & 12.57 & 34.8\%\\
 \hline
 MixMatch & 250 & 75 & 0.75 & 4e-5 & 0.0 & -\\
 \hline
 MixMatch + $\epsilon$mu & 250 & 75 & 1.0 & 4e-5 & 12.81 & 35.5\%\\
 \hline
 MixMatch & 500 & 75 & 0.75 & 4e-5 & 0.0 & -\\
 \hline
 MixMatch + $\epsilon$mu & 500 & 75 & 1.0 & 4e-5 & 12.62 & 35.0\%\\
 \hline
\end{tabular}

\label{table:appendixmixmatchmodelsettings_cifar10}
\end{table*}

\begin{table*}
\caption{This table summarizes the best model hyperparameter values for MixMatch and MixMatch + $\epsilon$mu on the SVHN dataset}
\centering
\normalsize

\begin{tabular}{ |p{2.2cm}|c||p{.5cm}|p{.5cm}|c|c|p{1.9cm}| }
 \hline
 \multicolumn{7}{|c|}{SVHN Best Model Hyperparameter Settings} \\
 \hline
 \multirow{2}{*}[-.9em]{Method} & \multirow{2}{*}[-.9em]{Labels} & \multirow{2}{*}[-.9em]{$w_S$} & \multirow{2}{*}[-.9em]{$\beta$} & \multirow{2}{*}[-.9em]{Weight Decay} & \multicolumn{2}{|c|}{Avg. learned $\epsilon$}\\
 \cline{6-7}
 & & & & & Magnitude & \% of Avg. Inter-Image Distance\\
 \hline
 MixMatch & 40 & 250 & 0.75 & 4e-5 & 0.0 & -\\
 \hline
 MixMatch + $\epsilon$mu & 40 & 250 & 1.0 & 4e-5 & 5.48 & 19.8\%\\
 \hline
 MixMatch & 250 & 250 & 0.75 & 4e-5 & 0.0 & -\\
 \hline
 MixMatch + $\epsilon$mu & 250 & 250 & 1.0 & 4e-5 & 4.98 & 18.0\%\\
 \hline
 MixMatch & 500 & 250 & 0.75 & 4e-5 & 0.0 & -\\
 \hline
 MixMatch + $\epsilon$mu & 500 & 250 & 1.0 & 4e-5 & 4.93 & 17.8\%\\
 \hline
\end{tabular}

\label{table:appendixmixmatchmodelsettings_svhn}
\end{table*}

\newpage


\bibliography{wileyNJD-APA.bib}%


\end{document}